\documentclass[journal, twoside]{IEEEtran}

\usepackage[utf8]{inputenc}
\usepackage[T1]{fontenc}
\usepackage{hyperref}
\usepackage{url}
\usepackage{booktabs}
\usepackage{amsfonts}
\usepackage{nicefrac}
\usepackage{microtype}
\usepackage{xcolor}
\usepackage{siunitx}
\usepackage{graphicx}
\usepackage{subcaption}
\usepackage{wrapfig}
\usepackage{amsmath}
\usepackage{cite}

\graphicspath{{figures/}}

\newcommand{\eg}{e.g.}
\newcommand{\ie}{i.e.}

\renewcommand{\paragraph}[1]{\textbf{#1}}

\begin{document}

\title{Temporal Reasoning via Audio Question Answering} 

\author{Haytham~M.~Fayek 
        and~Justin~Johnson
\thanks{H.~M.~Fayek is with Facebook Reality Labs, Redmond, WA, USA. \newline E-mail: haythamfayek@fb.com.}
\thanks{J.~Johnson is with Facebook AI Research, Menlo Park, CA, USA. \newline E-mail: jcjohns@fb.com.}
\thanks{Manuscript under review.} 
}

\markboth{}
{Fayek and Johnson: Temporal Reasoning via Audio Question Answering}

\maketitle

\begin{abstract}
Multimodal question answering tasks can be used as proxy tasks to study systems that can perceive and reason about the world.
Answering questions about different types of input modalities stresses different aspects of reasoning such as visual reasoning, reading comprehension, story understanding, or navigation.
In this paper, we use the task of Audio Question Answering (AQA) to study the temporal reasoning abilities of machine learning models.
To this end, we introduce the Diagnostic Audio Question Answering (DAQA) dataset comprising audio sequences of natural sound events and programmatically generated questions and answers that probe various aspects of temporal reasoning.
We adapt several recent state-of-the-art methods for visual question answering to the AQA task, and use DAQA to demonstrate that they perform poorly on questions that require in-depth temporal reasoning.
Finally, we propose a new model, Multiple Auxiliary Controllers for Linear Modulation (MALiMo) that extends the recent Feature-wise Linear Modulation (FiLM) model and significantly improves its temporal reasoning capabilities. 
We envisage DAQA to foster research on AQA and temporal reasoning and MALiMo a step towards models for AQA.
\end{abstract}



\section{Introduction}

\IEEEPARstart{A}{} central goal of artificial intelligence research has been to build systems that can perceive and reason about the world. In recent years, question answering has been used as a proxy task for this goal, where systems must answer natural language questions about another input such as an image~\cite{Antol2015,Johnson2017,zhu2016visual7w}, a piece of text~\cite{rajpurkar2016squad,weston2016towards}, an interactive environment~\cite{Das2018}, or a movie~\cite{Tapaswi2016}. Answering questions about different types of inputs stresses different aspects of reasoning such as visual reasoning, reading comprehension, story understanding, or navigation.

The world is dynamic, with events unfolding after each other over time. Systems that interact with the world must therefore be adept at \textit{temporal reasoning}. However the input modalities used by prior question answering benchmarks are ill-suited for isolating temporal aspects of reasoning. Images offer only a static snapshot of the world. Text may describe temporal sequences, but lacks intrinsic temporality. Navigating environments requires sequences of actions, but recent benchmarks navigate only static worlds. Movies are fundamentally temporal, but processing hours-long videos introduces computational barriers to studying temporal reasoning.

We believe that \textit{audio} is an appealing modality for investigating temporal reasoning, as it is fundamentally temporal and requires no spatial reasoning. In this paper, we therefore investigate the task of \textit{Audio Question Answering} (AQA) where systems must answer natural language questions pertaining to audio clips. In addition to being a testbed for temporal reasoning, advances in AQA have the potential to assist hearing-impaired individuals, much as advances in visual question answering could be used to assist visually-impaired individuals~\cite{bigham2010vizwiz,Gurari2018,lasecki2013answering}.

\begin{figure*}[t]
  \centering
  \begin{minipage}{0.53\textwidth}
  \includegraphics[width=\textwidth]{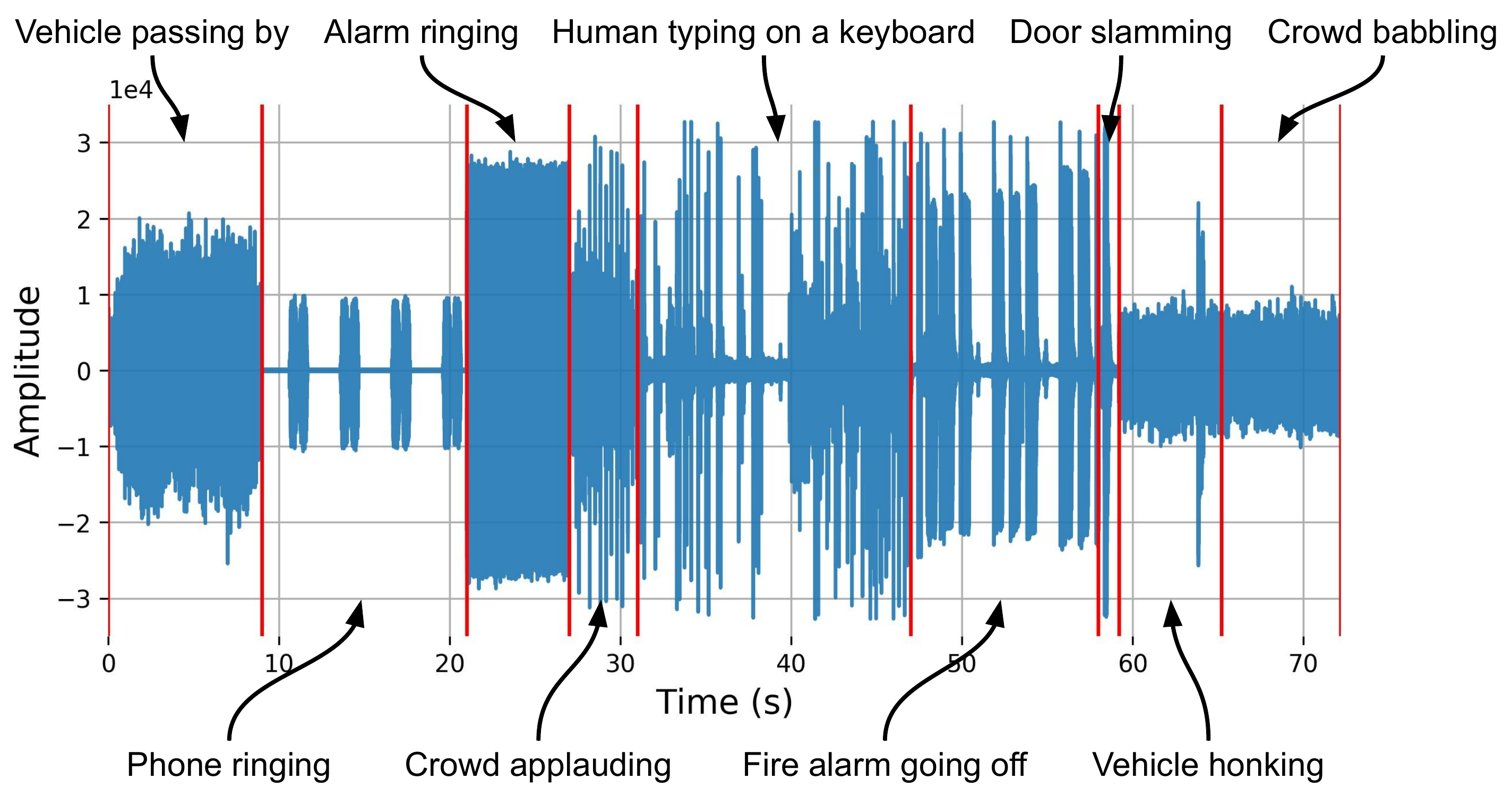}
  \end{minipage}%
  \hspace{0.2cm}%
  \begin{minipage}{0.45\textwidth}
      \footnotesize
      \begin{tabular}{@{}p{7.9cm}}
        Did you listen to any driver honking before the crowd babbling? \\
        \textbf{yes} \\
        \midrule
        Were the fourth and seventh sound events the same? \\
        \textbf{no} \\
        \midrule
        What was the shortest sound? \\
        \textbf{door slamming} \\
        \midrule
        What did you hear immediately after the crowd applauding? \\
        \textbf{human typing on a keyboard} \\
        \midrule
        How many times did you hear a vehicle passing by? \\
        \textbf{one} \\
        \midrule
        Was the first sound louder than the crowd babbling? \\
        \textbf{yes} \\
      \end{tabular}
  \end{minipage}%
  \caption{
    DAQA provides audio clips composed of natural sound events (left) and programmatically generated questions and answer about those audio clips that test various aspects of temporal reasoning (right). Note that the annotations on the audio clip are for illustration purposes only.
  } 
  \label{fig:aqa}
\end{figure*}

Towards these goals, this paper introduces the Diagnostic Audio Question Answering (DAQA) dataset for studying temporal reasoning in a controlled manner. DAQA provides \( 100,000 \) audio sequences ranging in length from approximately \( 10 \) seconds to approximately \( 3 \) minutes, constructed from a library of \( 400 \) natural sound events, and approximately \( 600,000 \) programmatically generated questions and answers that probe various aspects of temporal reasoning; see Figure~\ref{fig:aqa} for examples. Inspired by synthetic reasoning benchmarks for other domains~\cite{barrett2018measuring,Johnson2017,weston2016towards,yang2018dataset}, the design of DAQA minimizes bias that can be exploited by machine learning models to correctly answer questions without examining the question or the audio, and allows for fine-grained targeting of specific temporal reasoning skills.

We use DAQA to investigate the temporal reasoning abilities of several baseline models as well as adaptations of several recent state-of-the-art methods for visual question answering~\cite{Yang2016,Perez2018}. We demonstrate that although these methods show strong performance for visual question answering, they perform comparatively poorly on questions requiring in-depth temporal reasoning.

As a step towards making methods more adept at temporal reasoning, we introduce \textbf{M}ultiple \textbf{A}uxiliary Controllers for \textbf{Li}near \textbf{Mo}dulation (MALiMo), which extends and enhances the Feature-wise Linear Modulation (FiLM) model~\cite{Perez2018}. FiLM uses an auxiliary controller to process a supplementary input, \eg~the question, and predict parameters to modulate processing in a convolutional network that ingests the principal input, \eg~the image or the audio, to predict an answer. MALiMo augments FiLM with an additional auxiliary controller that receives subsampled features of the principal input and predicts parameters that modulate subsequent stages of processing in the convolutional network. This allows processing in the convolutional network to be altered as a function of both the principal and supplementary inputs, which facilitates relational and temporal reasoning. Experiments on DAQA show that MALiMo significantly outperforms other competing approaches, and that it leads to the largest improvements on questions that require in-depth temporal reasoning.

Code to generate the DAQA dataset and our models for the AQA task are publicly available%
\footnote{The code to generate the DAQA dataset and our models for the AQA task are available at \url{https://github.com/facebookresearch/daqa}.}.
We hope that this will foster research both on AQA and broadly on temporal reasoning.

\paragraph{Outline.}
The outline of the paper is as follows.
Section~\ref{sec:related} briefly reviews related work.
Section~\ref{sec:daqa} delineates the design and analysis of the DAQA dataset.
Section~\ref{sec:baselines} outlines oracle benchmarks and baseline models devised to divulge bias and assess the difficulty of the DAQA dataset.
Section~\ref{sec:model} presents our novel MALiMo model.
Section~\ref{sec:discussion} discusses the DAQA dataset and the MALiMo model as well as avenues for future work.
Section~\ref{sec:conclusion} concludes the paper.

\section{Related Work}%
\label{sec:related}

\paragraph{Question Answering Datasets.}
The AQA task and the DAQA dataset draw inspiration from datasets of natural language questions and answers about different input modalities, such as SQuAD~\cite{rajpurkar2016squad, rajpurkar2018know} for text; VQA~\cite{Antol2015,Goyal2017}, VizWiz~\cite{Gurari2018}, Visual Genome~\cite{krishna2017visual}, DAQUAR~\cite{malinowski2014multi}, and Visual7W~\cite{zhu2016visual7w} for images; TVQA~\cite{Lei2018}, MovieQA~\cite{Tapaswi2016}, Video-QA~\cite{zeng2017leveraging}, and~\cite{zhu2017uncovering} for videos; and EQA~\cite{Das2018} for embodied environments.

\paragraph{Diagnostic Reasoning Datasets.}
DAQA is synthetic and programmatically generated, allowing us
to control bias in the dataset and assess fine-grained temporal reasoning skills of AQA models. We draw inspiration from synthetic datasets for diagnosing other reasoning skills, such as bAbI~\cite{weston2016towards} for reading comprehension, CLEVR~\cite{Johnson2017}, COG~\cite{yang2018dataset}, and GQA~\cite{hudson2018gqa} for visual reasoning, and PGM~\cite{barrett2018measuring} for abstract reasoning.

\paragraph{Audio Question Answering.}
There has been some work on answering questions about images using spoken rather than written questions~\cite{zhang2017speech}, and answers to some questions about videos might require jointly reasoning about visual and audio cues~\cite{Lei2018,Tapaswi2016,zeng2017leveraging,zhu2017uncovering}. Most related to our work is the concurrent CLEAR dataset~\cite{Abdelnour2018}, which similar to DAQA, is a synthetic dataset of audio sequences and questions. All audio sequences in CLEAR are of fixed length and consist of \( 10 \) musical notes, while DAQA provides variable-length audio sequences with a variable number of more general audio events. CLEAR also adapts the question templates directly from the synthetic visual question answering dataset, CLEVR~\cite{Johnson2017}, while DAQA uses question types custom-built to emphasize temporal reasoning. We believe that these differences make DAQA a significantly more challenging dataset; \cite{Abdelnour2018} achieves \( 89.97\% \) accuracy on CLEAR with a FiLM~\cite{Perez2018} baseline, while our best and significantly improved FiLM model on DAQA only achieves \( 78.33\% \) accuracy.

\paragraph{Question Answering Models.}
There is a plethora of work on answering questions of various modalities, and a full survey is beyond the scope of this paper. Our work draws most directly from recent methods for visual question answering~\cite{anderson2018bottom,fukui2016multimodal,malinowski2015ask,singh2019pythia,Yang2016} and visual reasoning~\cite{hudson2018compositional,Johnson2017a,Perez2018,santoro2017simple}. Our MALiMo model is most closely related to FiLM~\cite{Perez2018} and conditional batch normalization~\cite{De2017}, though its auxiliary audio controller is related to transformers~\cite{Vaswani2017}, non-local networks~\cite{Wang2018}, and SENets~\cite{hu2018squeeze}.

\section{The Diagnostic Audio Question Answering (DAQA) Dataset}%
\label{sec:daqa}

The DAQA dataset is a synthetically generated dataset of <audio, question, answer> triplets.
The dataset requires temporal reasoning capabilities to identify, locate, compare, and count relevant audio events in a sequence of audio events, in order to correctly provide an answer to the posed question.
The DAQA dataset is envisaged to be useful for understanding, assessing, and benchmarking models for temporal reasoning in an AQA framework.

Each question in DAQA pertains to an \textit{audio clip} composed of several atomic \textit{audio events} arranged in sequence. We draw from a library of \( 400 \) audio events organized into \( 20 \) \textit{event types}. Each question is instantiated from one of \( 54 \) \textit{question templates} that test different aspects of temporal reasoning. Instantiating questions from templates involves several heuristics to reduce dataset bias. The answer to each question is one of \( 36 \) distinct values, so answering DAQA questions can be viewed as a classification problem. Overall, DAQA provides \( 100,000 \) audio clips and \( 599,294 \) questions and answers divided into standard training, validation, and test splits with \( 80,000 \); \( 10,000 \); and \( 10,000 \) audio clips and \( 399,924 \); \( 99,702 \); and \( 99,668 \) questions and answers each.

\begin{figure}[t]
    \centering
    \includegraphics[width=\linewidth]{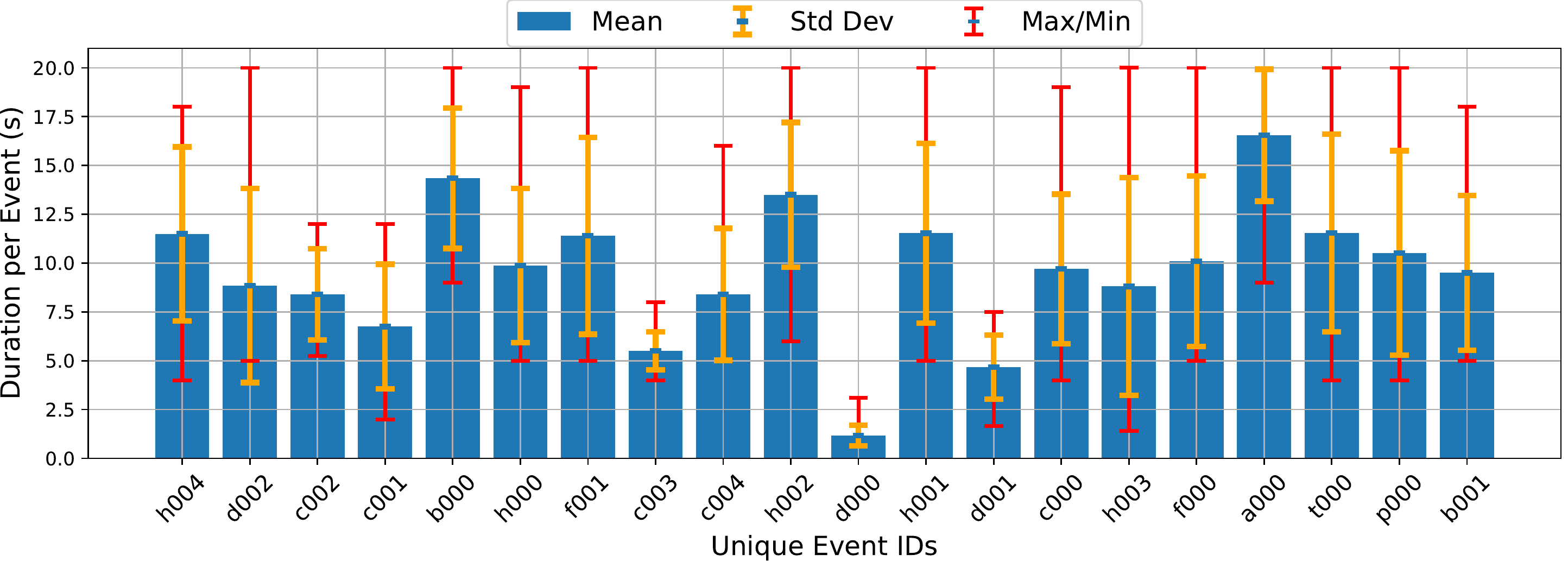} \\*
    \includegraphics[width=\linewidth]{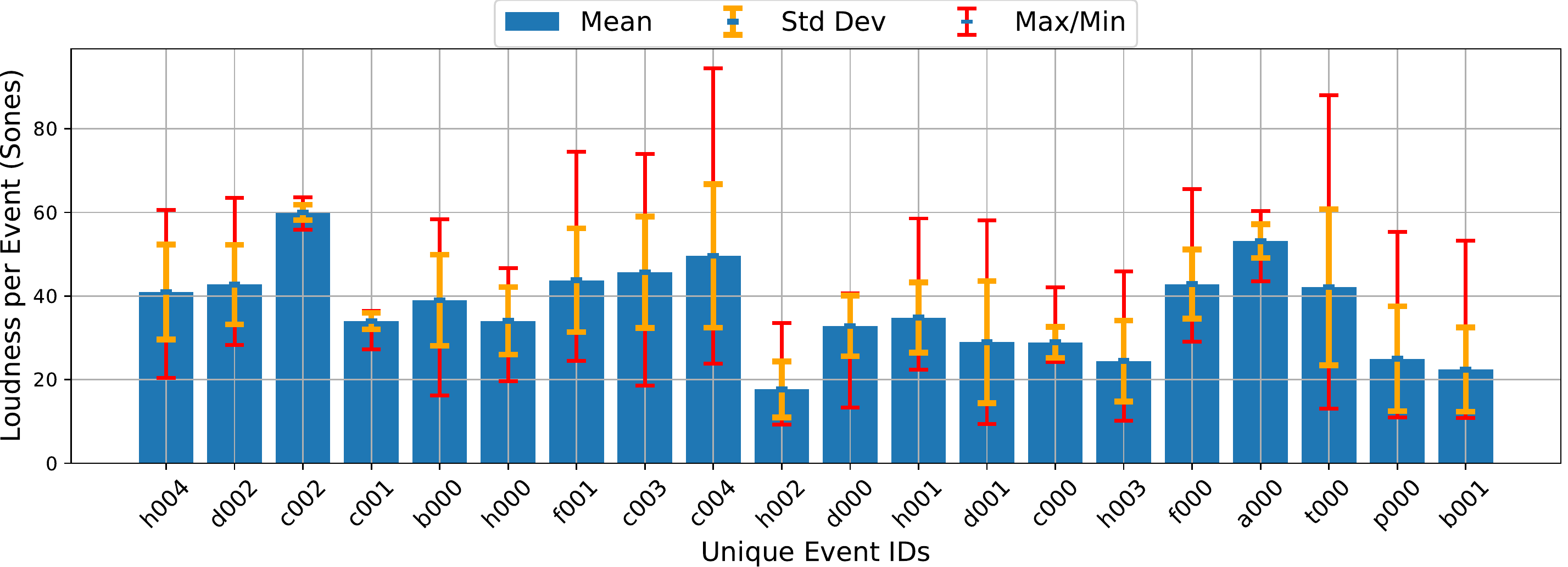}
    \caption{
    Mean, standard deviation, and limits of the duration (seconds) of the
      \( 20 \) instances for each of the \( 20 \) audio events (top).
    Mean, standard deviation, and limits of the loudness (sones) of the \( 20 \)
      instances for each of the \( 20 \) audio events (bottom). 
    }
    \label{fig:audioevents}
\end{figure}

\paragraph{Audio Events.}
Audio events are the atoms from which our audio clips are composed.
We use \( 20 \) different \textit{event types}\footnotemark\ chosen to span a wide variety of audio sources, locations, and sound types.
Each event type consists of a \textit{source} and an \textit{action} (\eg~\textit{crowd applauding}, \textit{dog barking}).
DAQA provides \( 20 \) unique instances of each event type, for a total of \( 400 \) unique audio event instances.
\( 91 \) of these events were recorded by the authors, while the other \( 309 \) were manually curated
from the balanced training subset of AudioSet~\cite{Gemmeke2017}.
As shown in Figure~\ref{fig:audioevents}, our \( 400 \) audio events vary significantly in
duration and loudness\footnotemark, both within each event type and across different event types.

\addtocounter{footnote}{-1}
\footnotetext{%
      \label{note:id}Events are referred to by a unique ID as follows;
      a000: \textit{aircraft flying over},
      b000: \textit{band playing},
      b001: \textit{bird singing},
      c000: \textit{crowd babbling},
      c001: \textit{crowd applauding},
      c002: \textit{crowd rioting},
      c003: \textit{car honking},
      c004: \textit{car passing by},
      d000: \textit{door slamming},
      d001: \textit{doorbell ringing},
      d002: \textit{dog barking},
      f000: \textit{fire engine passing by},
      f001: \textit{fire alarm going off},
      h000: \textit{human speaking},
      h001: \textit{human laughing},
      h002: \textit{human typing on a keyboard},
      h003: \textit{human whistling},
      h004: \textit{human operating a machine},
      p000: \textit{phone ringing},
      and t000: \textit{storm thundering}.}
\stepcounter{footnote}
\footnotetext{We compute loudness following the time-varying loudness model~\cite{glasberg2002model} using the Genesis Loudness Toolbox.}

We divide our \( 20 \) event types into \( 5 \) \textit{discrete} types (\textit{aircraft flying over}, \textit{car passing by}, \textit{door slamming}, \textit{human speaking}, and \textit{human laughing}) and \( 15 \) \textit{continuous} types (all others).
A sequence of discrete events of the same type (\eg~two consecutive \textit{door slamming} events) can be easily segmented by a listener into disjoint events, while sequential continuous events of the same type (\eg~two consecutive \textit{crowd babbling} events) would likely be heard as a single continuous event.

\paragraph{Audio Clips.}
Audio clips are generated by concatenating between \( 5 \) and \( 12 \) random audio events.
We only allow consecutive \textit{discrete} events of the same type; clips with adjacent \textit{continuous} events of the same type are rejected.
We randomly overlap successive events by up to \SI{500}{ms}, and add normally distributed background noise to half of our audio clips.
We generated \( 100{,}000 \) audio clips, divided into training, validation, and test splits of \( 80{,}000 \); \( 10{,}000 \); and \( 10{,}000 \) clips respectively.
We ensured that no validation or test clips have the exact same sequence of events instances as any training clip.

Since each audio clip contains a variable number of audio events and each audio event varies in length, our generated audio clips vary significantly in length: clips in the training set vary from \( 10.5 \) to \( 178.2 \) seconds, with a mean and standard deviation of \( 80.8 \pm 26.3 \) seconds.
We believe that this wide variety in clip lengths is necessary for evaluating temporal reasoning.

Each audio clip is annotated with the order, identity, duration, and loudness of its constituent events.
We use these annotations only for analysis and oracle baseline models; in particular our AQA models are trained end-to-end for question answering and are not supervised with this information.

\begin{figure}[t]
    \centering
    \includegraphics[trim={35px 20px 20px 20px},clip,width=\linewidth]{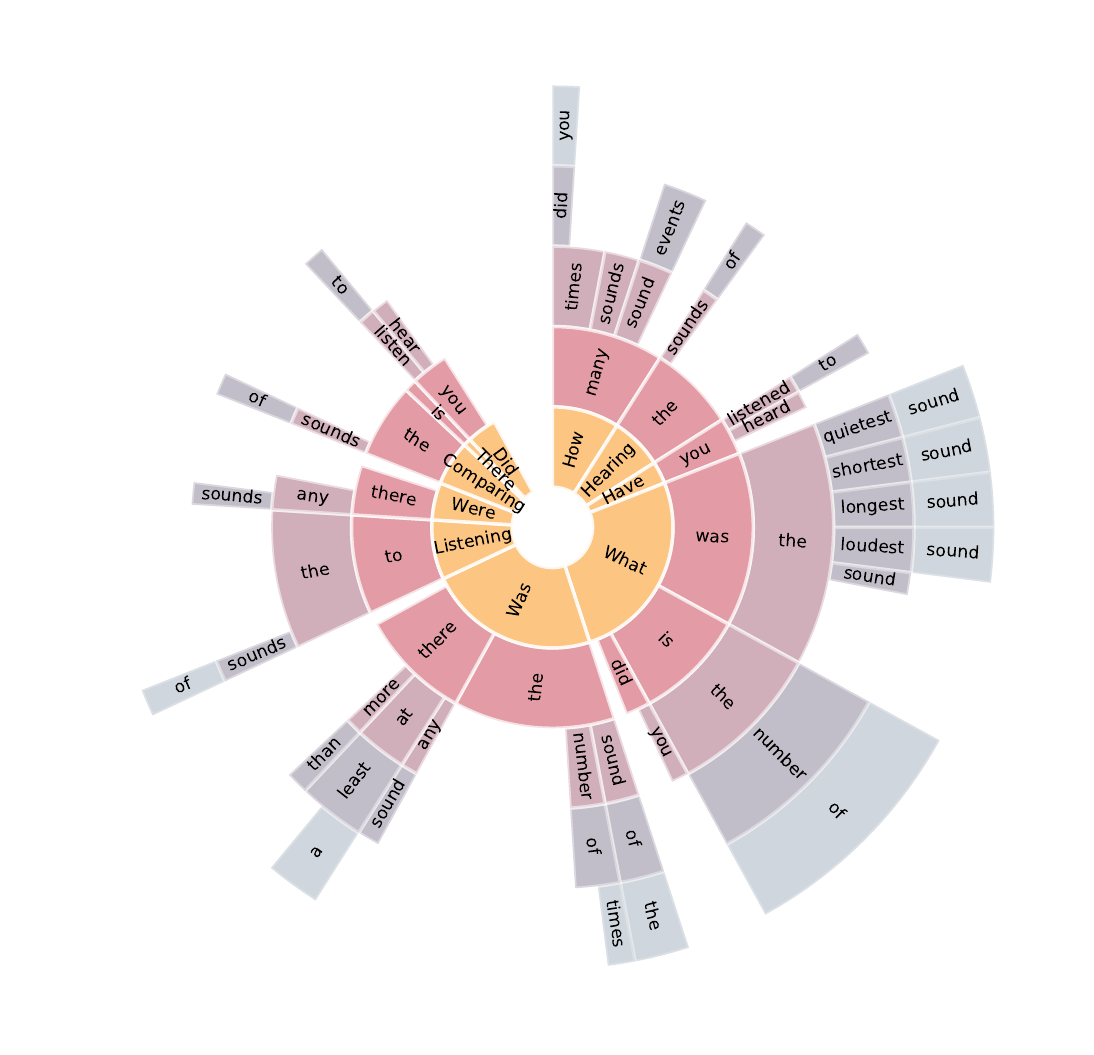}
    \caption{Distribution of the first five words for all questions in the training
      set of the DAQA dataset. The innermost ring represents the first words and
      radiating rings represent subsequent words. Arc lengths are proportional to the number of
      questions with the word. Words accounting for less than \( 1\% \) were omitted for clarity.}
    \label{fig:questions}
\end{figure}

\paragraph{Question Templates.}
Questions are generated programmatically from one of \( 54 \) manually designed \textit{question templates},
Each template contains several \textit{placeholder} values of various types; given annotations for an audio clip, the template can be \textit{instantiated} by choosing values for each placeholder, giving rise to a (question, answer) pair for the clip.
For example, the template \texttt{What did you hear <RO> the <S> <A>?} has placeholders \texttt{<RO>} for a preposition (\textit{before} or \textit{after}), \texttt{<S>} and \texttt{<A>} for the source and action of an event type.
To increase linguistic diversity, each template provides several logically equivalent English phrasings, such as \texttt{What was the sound <RO> the <S> <A>?} for the template above; we also randomly replace some words with synonyms (\eg~\textit{person} for \textit{human}).
Associated with each template are short Python programs to generate or verify answers given placeholder values and audio clip annotations.

\begin{figure*}[t]
    \centering
    \includegraphics[width=0.9\textwidth]{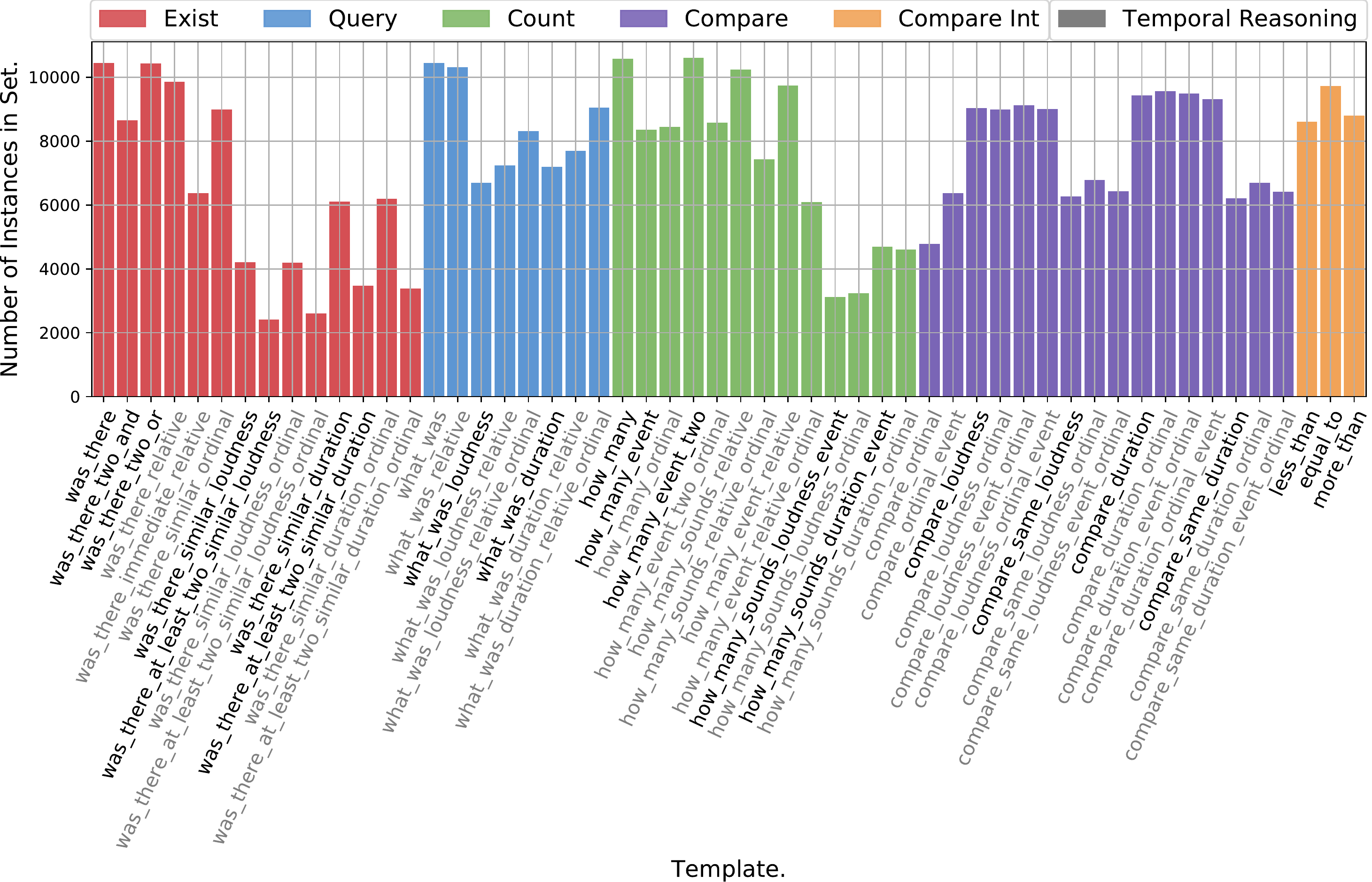}
    \caption{Number of examples per template in the training set of the DAQA dataset sorted by color with respect to reasoning skills. Templates that require temporal reasoning, \eg~contain prepositions: before / after, or ordinals, are {\color{gray}{highlighted}} in {\color{gray}{gray}}.}
    \label{fig:count}
\end{figure*}

To stress temporal reasoning, templates use a variety of mechanisms to refer to audio events.
Audio events can be referenced using event type (\textit{the alarm ringing}), absolute ordinal position (\textit{the third sound}), relative position (\textit{the sound immediately after the alarm ringing}), absolute duration or loudness, (\textit{the loudest sound}), or relative duration or loudness (\textit{sounds longer than the alarm ringing}).

Templates may also be grouped by reasoning skills: \textit{exist} templates ask whether an event is present (\textit{was there an alarm ringing?}), \textit{query} templates ask about event properties (\textit{What was the loudest sound?}), \textit{compare} templates require comparisons between two events (\textit{was the first sound louder than the vehicle honking?}), \textit{count} templates require counting events meeting some condition (\textit{how many events were louder than the alarm ringing?}), and \textit{compare integer} templates require comparing sizes of two event sets (\textit{were there more alarms ringing than sounds before than the door slamming?}).

\paragraph{Template Instantiation.}
Instantiating a template requires choosing a value for each of its placeholders.
This process involves several subtle issues that require special attention.

Some choices of placeholders give questions that are \textit{invalid} for some audio clips -- for example the question \textit{What was the sound before the human speaking?} is valid only for clips with exactly one \textit{human speaking} event. For some (template, clip) pairs, most possible template instantiations will be invalid, so randomly sampling placeholder values and rejecting invalid questions would be inefficient. We overcome this by incorporating logic into each template to sample only from valid placeholder values, and we terminate early if a placeholder has no valid values.

\begin{figure*}[t]
    \centering
    \includegraphics[width=0.9\textwidth]{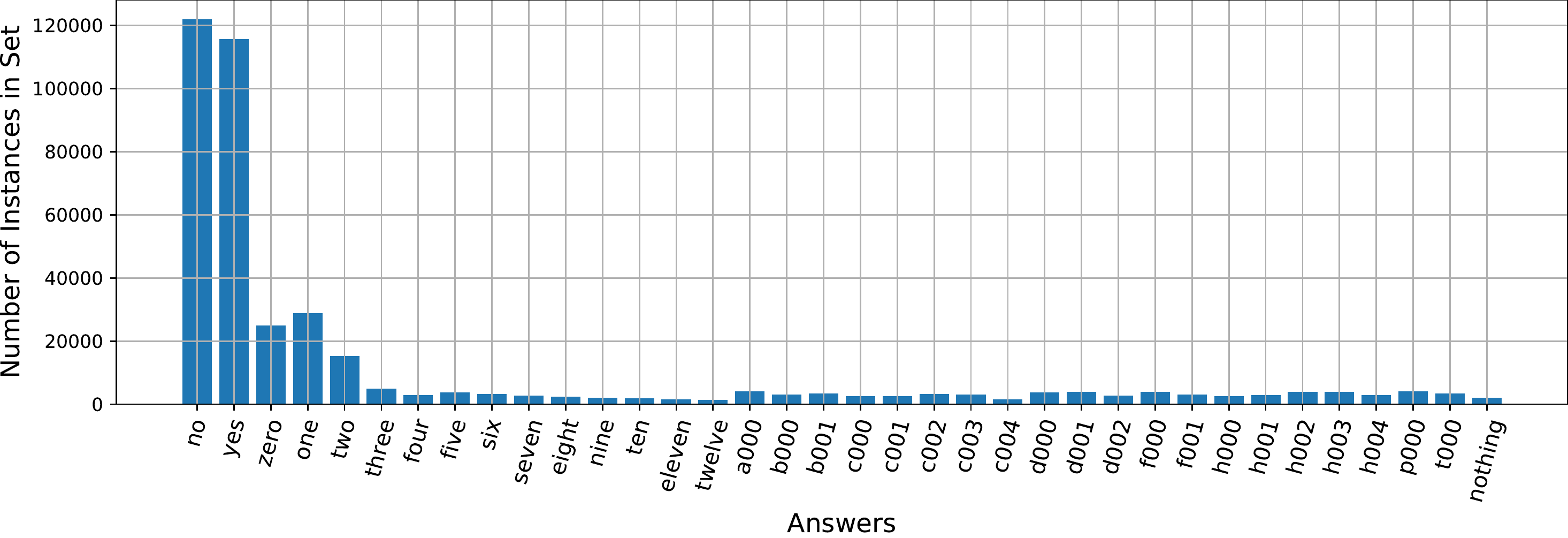}
    \caption{Frequency of answers in the training set of the DAQA dataset.}
    \label{fig:answers}
\end{figure*}

Sampling questions uniformly at random from all valid template instantiations can lead to severe answer imbalance. For example, consider questions of the form \textit{Was there a <S> <A> immediately after the door slamming?} -- as long as the clip contains a door slamming, any values for \texttt{<S>} and \texttt{<A>} give valid questions; however answers will be heavily biased towards \textit{No}, introducing a significant question-conditional bias and allowing models to achieve high accuracy for these questions without examining the audio clip. We overcome this problem heuristically by rejecting questions that lead to a large difference (\( >5\% \)) between the most and least common answers for each template. Figure~\ref{fig:answers} shows the frequency of each answer in the training set; the relatively uniform distribution within each answer type is a result of this heuristic.

We attempt to instantiate 5 questions for each training clip and 10 questions for each validation and test clip; however due to the above heuristics, we are unable to sample the desired number of questions for all clips. Our training set thus contains \( 80,000 \) clips and \( 399,924 \) questions (\( 172,126 \) of which are unique); the validation set has \( 10,000 \) clips and \( 99,702 \) questions, and the test set has \( 10,000 \) clips and \( 99,668 \) questions. Questions in the training set range in length from \( 5 \) to \( 27 \) words, with a mean length and standard deviation of \( 12.89\pm3.86 \) words.

\paragraph{Answers.}
Each questions's answer is one of \( 36 \) possible values: \textit{yes}, \textit{no}, the \( 20 \) event types, \textit{nothing}, and integers \( 0 \) to \( 12 \) (inclusive). Figure~\ref{fig:answers} shows the frequency of these answers on the training set.

\section{Benchmark Experiments}%
\label{sec:baselines}

We use several baselines as well as adaptations of recent state-of-the-art models for visual question answering to investigate and benchmark DAQA and the AQA task.

\paragraph{Experimental setup.}
The AQA task, \(Audio + Question \rightarrow Answer\), is formulated as a multi-class classification problem, and all models are trained to maximize the likelihood of the correct answer  \(p(Ans | A, Q)\). 
DAQA is split into training, validation, and test sets; see Section~\ref{sec:daqa}.
The validation set was used to design experiments and tune hyperparameters, such as the number and size of the layers, independently per model via grid search.
Each model was evaluated on the test set only once after finalizing all experimental details and hyperparameters.

Each audio clip was split into \SI{25}{ms} frames with a stride of \SI{10}{ms}, and a Hamming window was applied, then \( 64 \log \)-Mel-Frequency Spectral Coefficients (MFSCs) were extracted from each frame.
The mean and standard deviation were normalized per coefficient to zero and one respectively using the mean and standard deviation computed from the training set only.

All models were trained from scratch, i.e.\ no pre-trained word embeddings or models were used.
Neural models were trained via stochastic gradient descent with a mini-batch size of \( 40 \) using Adam~\cite{Kingma2015} with learning rate \( \alpha = 1 \times 10^{-4} \) and weight decay \( \lambda = 1 \times 10^{-5}\) for \( 100 \) epochs, and the validation set was used to select the best model during training.

The following is a detailed description of the baselines and models.
The results are listed in Table~\ref{table:results}.

\paragraph{Random / Mode Answer.}
We choose either a random or the most frequent (mode) answer from the training set.
We also evaluate per-template (P/T) versions which output either a random valid answer for each question's template, or the most frequent training set answer for each template.

\paragraph{Audio Only.}
This model ignores the question, and uses a Fully Convolutional Network (FCN) on the audio to predict the answer.
The FCN model has
five convolutional blocks similar to those in VGG~\cite{Simonyan2015}; each block comprises two convolution layers with Batch Normalization (BatchNorm) and Rectified Linear Unit (ReLU) nonlinearities, and a max pooling layer with a \( 2 \times 2 \) window with stride \( 2 \) after the second convolution layer;
followed by two standard convolution layers with BatchNorm and ReLUs, a final convolution layer, then global average pooling across each channel.
The convolution layers in the first convolutional block have \( 32 \) filters of size \( 3 \times 12 \) with stride \( 1 \times 9 \), whereas the number of filters double every convolutional block up to \( 512 \) filters, all of which are of size \(3 \times 3\) with stride \( 1 \), except the penultimate layer, which has \( 1024 \) filters of size \( 1 \times 1 \) and the final layer which has \( K = 36 \) filters of size \( 1 \times 1 \), where \( K \) corresponds to the number of classes in DAQA.

We train models both for the entire dataset as well as separate models per-template (P/T).
The overall accuracy of the audio only model for all templates is comparable to the Mode Answer baseline (\( 31.27\% \) vs \( 30.52\% \)).
The overall accuracy of the audio only per-template models is slightly better than the Mode Answer P/T baseline (\( 39.72\% \) vs \( 43.59\% \)).
These results indicate that DAQA has minimal audio-conditional bias.

\paragraph{Question Only.}
We use several baselines that ignore the audio.
We train logistic regression models on a one-hot encoding of each question's template (Question Template) and on Bag of Words encodings of questions (Q-only Logit).
We also train models that encode the questions with \( 128 \)-dimensional word embeddings followed by a two-layer unidirectional Long Short-Term Memory (LSTM) network with \( 512 \) units per layer and predict answers with a linear projection of the final LSTM hidden state (Q-Only LSTM).

We train the latter two models both for the entire dataset as well as separate models per-template (P/T).
Per-template LSTMs are not substantially better than the Mode Answer P/T baseline (\( 44.93 \% \) vs \( 39.72\% \)), suggesting that DAQA has minimal question-conditional bias.

\begin{table*}[t]
  \renewcommand{\arraystretch}{1.1}
  \caption{
    Audio question answering performance on the DAQA test set.
    We use baselines to divulge bias in the dataset and adapt recently proposed models for visual question answering~\cite{Perez2018,Yang2016} to the AQA task.
    Our proposed MALiMo model outperforms prior methods due to improved temporal reasoning capabilities.
  }%
  \label{table:results}
  \centering
  \begin{tabular}{l | c | ccccc | c}
    \toprule
    Model                                             & \# Params & Exist & Query & Count & Compare & Compare Int & All (\%) \\
		\midrule
		Random Answer    																	& -- & 2.55 & 2.96 & 2.73 & 2.80 & 2.88 & 2.76 \\
		Mode Answer     													        & -- & 51.50 & 0.00 & 0.00 & 51.33 & 50.69 & 30.52 \\
		Random Answer P/T												          & -- & 49.56 & 4.75 & 15.66 & 50.10 & 49.59 & 34.15 \\
		Mode Answer P/T									                  & -- & 52.02 & 7.52 & 32.54 & 51.41 & 52.05 & 39.72 \\
    \midrule
		Audio Only (FCN)																	& 7.65M & 55.64 & 0.03 & 0.0 & 50.66 & 51.54 & 31.27 \\
		Audio Only P/T (FCN)														  & -- & 55.90 & 17.52 & 37.62 & 50.73 & 57.24 & 43.59 \\
		\midrule
    Question Template 																& -- & 52.02 & 7.48 & 32.53 & 51.41 & 52.05 & 39.72 \\
    Q-Only (Logit)								                    & -- & 53.02 & 12.88 & 33.98 & 51.46 & 51.95 & 41.19 \\
    Q-Only P/T (Logit)              	                & -- & 53.75 & 12.38 & 34.56 & 51.95 & 52.37 & 41.59 \\
		Q-Only (LSTM)															        & 3.45M & 54.20 & 13.09 & 34.22 & 61.23 & 51.04 & 44.53 \\
		Q-Only P/T (LSTM)								                  & -- & 53.87 & 13.48 & 34.78 & 61.72 & 52.93 & 44.93 \\
		\midrule
		FCN-LSTM									 												& 9.79M & 67.88 & 30.07 & 46.42 & 59.99 & 62.99 & 53.65 \\
    ConvLSTM-LSTM									 										& 14.78M & 71.52 & 45.72 & 52.09 & 62.13 & 65.11 & 59.22 \\
		FCN-LSTM-SA 				                              & 10.58M & 75.52 & 45.68 & 57.1 & 61.67 & 70.5 & 61.48 \\
    \midrule
    FiLM-512 (2 FiLM layers)                          & 5.49M & 77.27 & 56.23 & 57.77 & 62.17 & 75.22 & 64.27 \\
    FiLM-512 (4 FiLM layers)                          & 6.35M & 77.63 & 56.69 & 59.11 & 62.23 & 76.53 & 64.84 \\
    FiLM-512 (6 FiLM layers)                          & 7.21M & 79.6 & 60.54 & 62.61 & 63.56 & 81.42 & 67.49 \\
    FiLM-512 (8 FiLM layers)                          & 8.07M & 80.08 & 72.91 & 77.22 & 65.93 & 89.23 & 74.39 \\
    FiLM-512 (10 FiLM layers)                         & 8.93M & 81.03 & 76.37 & 75.28 & 66.63 & 94.59 & 75.3 \\
    FiLM-512 (12 FiLM layers)                         & 9.79M & 82.92 & 80.13 & 79.74 & 69.83 & 93.68 & 78.33 \\
    \midrule
    FiLM-1024 (2 FiLM layers)                         & 15.73M & 78.33 & 56.57 & 59.06 & 62.61 & 76.7 & 65.09 \\
    FiLM-1024 (4 FiLM layers)                         & 16.85M & 79.29 & 61.04 & 63.54 & 62.92 & 84.87 & 67.76 \\
    FiLM-1024 (6 FiLM layers)                         & 17.97M & 79.76 & 61.22 & 64.77 & 63.76 & 88.17 & 68.66 \\
    FiLM-1024 (8 FiLM layers)                         & 19.09M & 79.69 & 75.86 & 70.15 & 63.93 & 93.53 & 72.79 \\
    FiLM-1024 (10 FiLM layers)                        & 20.21M & 79.98 & 61.33 & 64.35 & 63.4 & 85.48 & 68.34 \\
    FiLM-1024 (12 FiLM layers)                        & \textbf{21.33M} & 78.77 & 76.46 & 75.28 & 64.96 & 88.13 & 73.87 \\
    \midrule
    MALiMo (1 Block; ours)                            & 8.91M & 78.34 & 46.73 & 64.18 & 62.53 & 83.45 & 65.08 \\
    MALiMo (2 Blocks; ours)                           & 9.77M & 80.86 & 78.21 & 71.52 & 67.34 & 90.75 & 74.65 \\
    MALiMo (3 Blocks; ours)                           & 10.63M & 82.62 & 81.09 & 77.22 & 68.95 & 91.16 & 77.39 \\
    MALiMo (4 Blocks; ours)                           & 11.49M & 85.69 & 86.33 & 80.49 & 74.93 & 95.64 & 81.87 \\
    MALiMo (5 Blocks; ours)                           & 12.34M & 89.10 & 88.80 & 85.05 & 79.63 & 95.84 & 85.59 \\
    \textbf{MALiMo} (6 Blocks; ours)                  & 13.20M & \textbf{91.08} & \textbf{90.60} & \textbf{86.35} & \textbf{86.53} & \textbf{97.04} & \textbf{88.86} \\
    \bottomrule
  \end{tabular}
\end{table*}

\paragraph{FCN-LSTM.}
This model is composed of an FCN, identical to the FCN used for the audio only model, to encode the audio and a two-layered LSTM, with \( 512 \) units in each layer, to encode \( 256 \)-dimensional word embeddings of each word in the question.
Both representations are concatenated and fed to a fully connected neural network with a single hidden layer of \( 1024 \) units and ReLUs to predict an answer.

This model performs relatively poorly (\( 53.65\% \)) but outperforms prior audio only and question only baselines.

\paragraph{ConvLSTM-LSTM.}
This model is composed of five convolutional blocks identical to the VGG blocks in the audio only model, that encode the audio into a variable-sized representation, which is fed to a single-layered LSTM, with \( 512 \) units, that encodes the variable-sized representation into a fixed-sized representation of the audio; another two-layered LSTM, with \( 512 \) units in each layer, is used to encode \( 256 \)-dimensional word embeddings of each word in the question into a fixed-sized representation of the question.
Both representations are concatenated and fed to a fully connected neural network with a single hidden layer of \( 1024 \) units and ReLUs to predict an answer.

This model outperforms all baselines and the FCN-LSTM model (\( 59.22\% \) vs \( 53.65\% \)).

\paragraph{FCN-LSTM-Stacked Attention.}
This model is similar to the FCN-LSTM model described above, albeit with two layers of Stacked Attention (SA)~\cite{Yang2016}.
Concretely, instead of concatenating the audio representations from the FCN and the encoded question representations from the LSTM, the global average pooling layer in the FCN is replaced with an adaptive average pooling layer with a \( 2 \times 8 \) output, and two SA layers are employed that ingest the representations from the FCN and LSTM to produce attention maps (see~\cite{Yang2016} for details), which are then added to the FCN representations, and fed to a fully connected neural network with a single hidden layer of \( 1024 \) units and ReLUs to predict an answer.

This model, augmented with SA, slightly outperforms the ConvLSTM-LSTM model (\( 61.48\% \) vs \( 59.22\% \)) using less parameters (\( 14.78 \)M vs \( 10.58 \)M).

\paragraph{FiLM.}
FiLM~\cite{Perez2018} adopts feature-wise linearly modulated layers conditioned on the question that manipulate the intermediate representations of a neural network to adaptively influence the output of the neural network by applying an affine transformation as follows. Let \( \mathbf{H}^{(l)} \in \mathbb{R}^{n \times h \times w} \) be the output of a linear convolution layer \( l \) that has \( n \) filters,
the modulation parameters are applied per feature map \( \mathbf{H}^{(l)}_i \):
\begin{equation}
  \label{eq:film}
  FiLM(\mathbf{H}^{(l)}_i | \boldsymbol{\gamma}^{(l)}_i,\boldsymbol{\beta}^{(l)}_i) = \boldsymbol{\gamma}^{(l)}_i \mathbf{H}^{(l)}_i + \boldsymbol{\beta}^{(l)}_i,
\end{equation}
where \( \mathbf{H}^{(l)}_i \in \mathbb{R}^{h \times w} \) is the feature map of filter \( i \) of convolution layer \( l \) and \( \boldsymbol{\gamma}^{(l)},\boldsymbol{\beta}^{(l)} \in \mathbb{R}^{n} \) are the modulation parameters from \( f_{c} \), \ie,
\( ( \boldsymbol{\gamma},\boldsymbol{\beta} ) = f_{c}(\mathbf{z}) \),
such that \( f_{c} \) is a neural network and \( \mathbf{z} \) is the question.

\begin{figure*}[t]
  \centering
  \includegraphics[width=\textwidth]{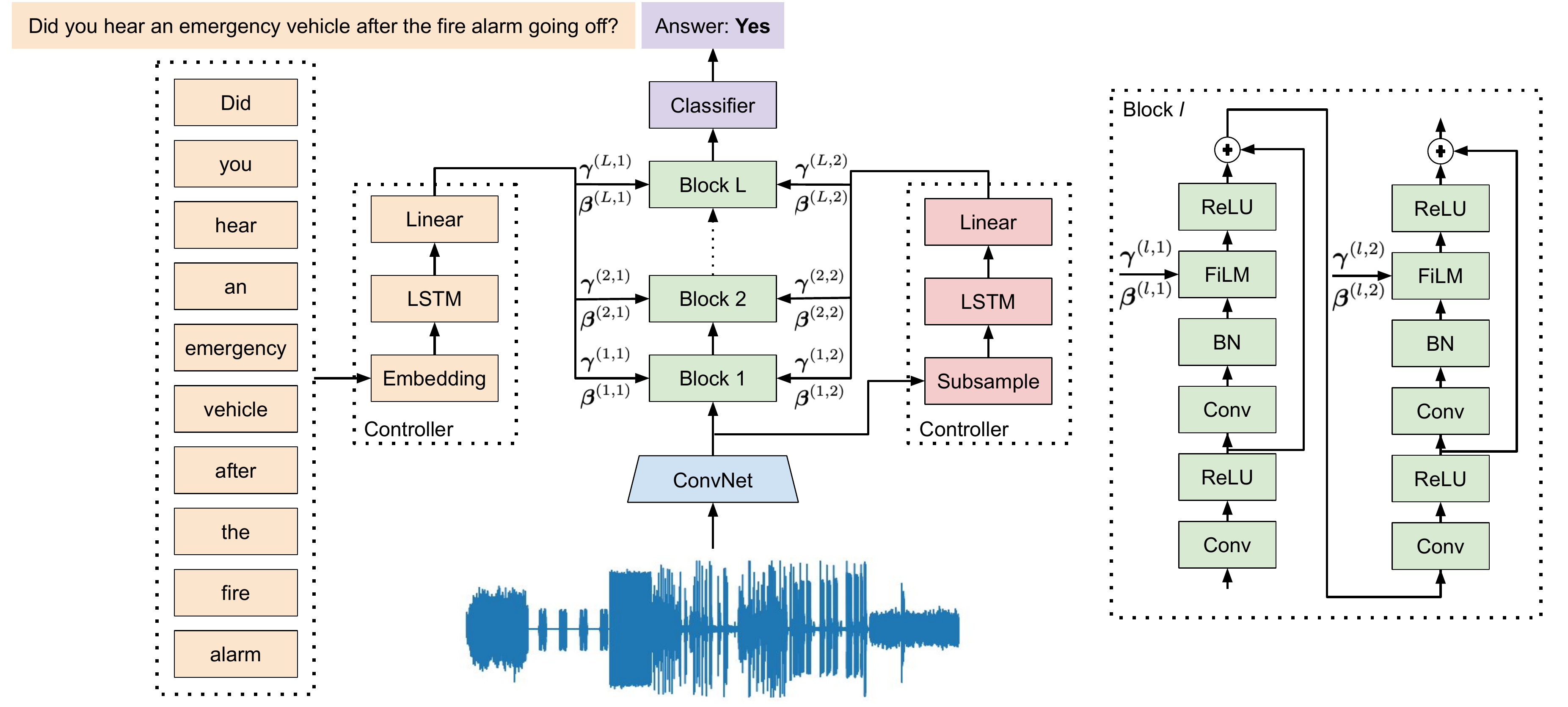}
  \caption{
    Multi Auxiliary Controllers for Linear Modulation (MALiMo) for AQA.
    Two \textit{auxiliary controllers}: one ingests the question text, and another operates on subsampled audio features, are used to modulate intermediate representations of a convolutional network that ingests the audio and predicts an answer.
  }%
  \label{fig:malimo}
\end{figure*}

We adapt the FiLM model to the AQA task.
Our FiLM model is composed of three convolutional blocks, identical to the VGG blocks in the audio only model, that are used to process the audio.
A two-layered LSTM, with \( 512 \) or \( 1024 \) units per layer, encodes the question and produces the modulation parameters \( ( \boldsymbol{\gamma}, \boldsymbol{\beta} ) \) for residual convolutional blocks that ingest the processed representations from the preceding block.
Each modulated block comprises a convolution layer with \( 128 \) \( 3 \times 3 \) filters and ReLUs, followed by another convolution layer with \( 128 \) \( 3 \times 3 \) filters, BatchNorm (with no affine transformation), feature-wise linear modulation, and ReLUs, and a residual connection from the output of the first ReLUs to output of the second ReLUs.
Two coordinate feature maps indicating relative time and frequency positions respectively, scaled from \( -1 \) to \( 1 \), were concatenated to the input in each modulated block.
A convolution layer with \( 512 \) \( 1 \times 1 \) filters followed by global average pooling across each channel is applied to the output of the final modulated block and the output is fed to a fully connected neural network with a single hidden layer of \( 1024 \) units with ReLUs to predict an answer.

\begin{figure*}[t]
  \centering
  \includegraphics[width=0.9\textwidth]{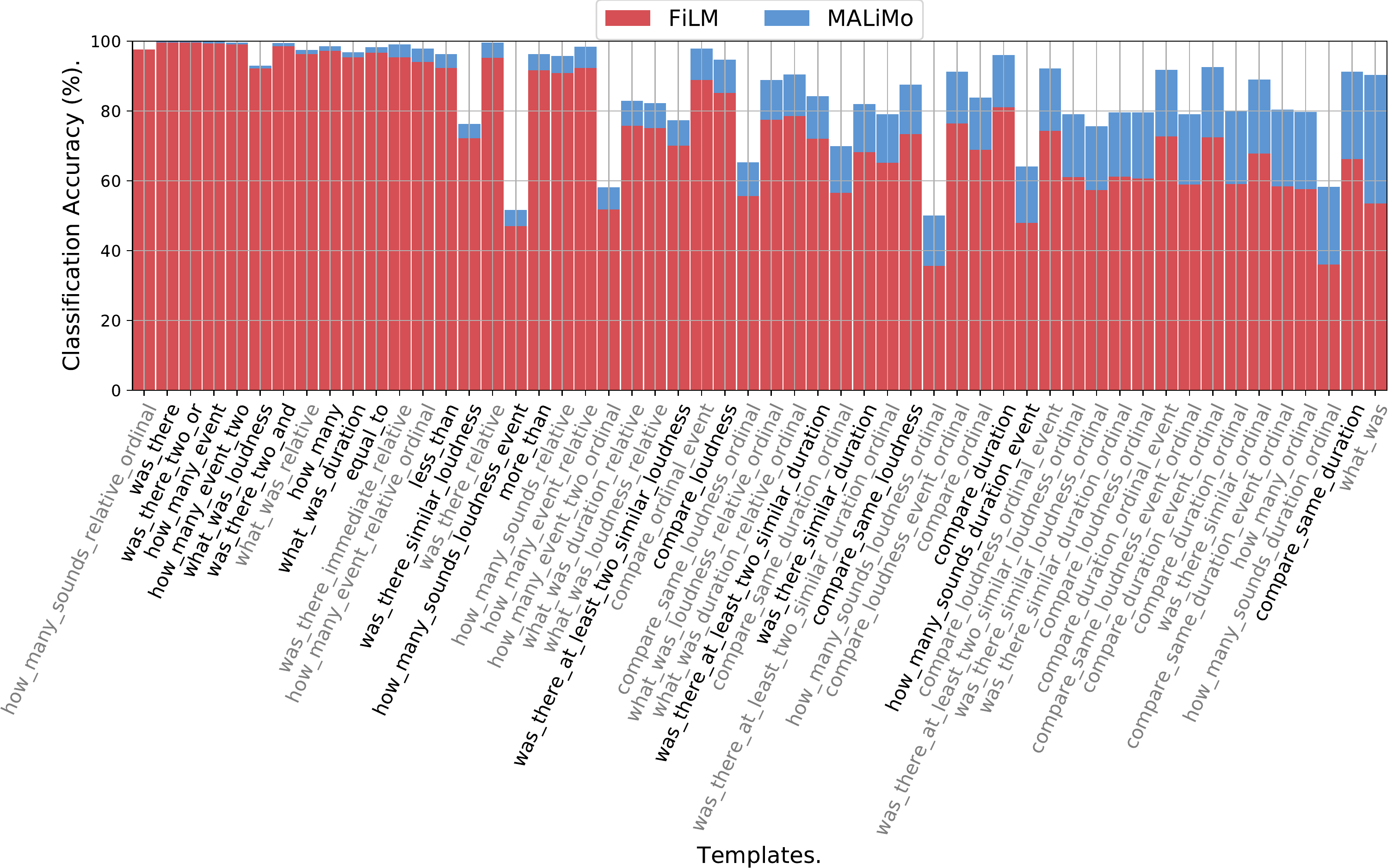}
  \caption{Audio question answering performance per template on the DAQA test set for FiLM (FiLM-512; 12 layers) and MALiMo (6 blocks).
  Templates {\color{gray}{highlighted}} in {\color{gray}{gray}} require temporal reasoning, \eg~contain prepositions: before / after, or ordinals.}%
  \label{fig:templates}
\end{figure*}

We evaluate FiLM exhaustively by varying the number of units in network \( f_{c} \) and the number of modulated blocks as shown in Table~\ref{table:results}.
FiLM outperforms all previous baselines and models.
The best FiLM model achieves an overall accuracy of \( 78.33\% \).
Nevertheless, most of the classification error in the best performing FiLM model stems from questions and templates that require temporal reasoning as illustrated in Figure~\ref{fig:templates} and discussed later in Section~\ref{sec:discussion}.

\section{Multi Auxiliary Controllers for Linear Modulation (MALiMo)}%
\label{sec:model}

MALiMo comprises a main module \( f_{x} \) and auxiliary controller(s) \( f_{c} \).
These can all be implemented as end-to-end differentiable neural networks.
The main module \( f_{x} \) maps the principal input \( \boldsymbol{x} \) to the output \( \boldsymbol{y} \).
Each auxiliary controller \( f^{(k)}_{c} \) ingests either the principal input  \( \boldsymbol{x} \) or another supplementary input \( \boldsymbol{z}^{(k)} \) and produces a set of parameters \( ( \boldsymbol{\gamma}^{(k)}, \boldsymbol{\beta}^{(k)} ) \) that modulate the intermediate representations in the main module \( f_{x} \), where \( k \) denotes the \( k^{th} \) controller.
This allows computation in the main module \( f_{x} \) to be conditioned on the outputs \( ( \boldsymbol{\gamma}, \boldsymbol{\beta} ) \) of the auxiliary controller(s) \( f_{c} \).
Crucially, at least one auxiliary controller acts on the principal input \( \boldsymbol{x} \), or a processed version of the principal input, which enables the auxiliary controller to alter the intermediate representations in the main module acting on the principal input as a function of the principal input itself to facilitate relational and temporal reasoning, emulating the role of self-attention~\cite{Vaswani2017} and non-local operators~\cite{Wang2018}.
The modulation parameters \( ( \boldsymbol{\gamma}, \boldsymbol{\beta} ) \) from the auxiliary controller(s) \( f_{c} \) are injected into the main module \( f_{x} \) as feature-wise linear transformations on the intermediate representations in a stacked manner in each modulated layer as follows.

Let \( \mathbf{H}^{(l)} \in \mathbb{R}^{n \times h \times w} \) be the output of a linear convolution layer \( l \) that has \( n \) filters,
the modulation parameters are applied per feature map \( \mathbf{H}^{(l)}_i \)~\cite{Perez2018}:
\begin{equation}
  \label{eq:malimo}
  MALiMo(\mathbf{H}^{(l)}_i | \boldsymbol{\gamma}^{(l,k)}_i,\boldsymbol{\beta}^{(l,k)}_i) = \boldsymbol{\gamma}^{(l,k)}_i \mathbf{H}^{(l)}_i + \boldsymbol{\beta}^{(l,k)}_i,
\end{equation}
where \( \mathbf{H}^{(l)}_i \in \mathbb{R}^{h \times w} \) is the feature map of filter \( i \) of convolution layer \( l \) and \( \boldsymbol{\gamma}^{(l,k)},\boldsymbol{\beta}^{(l,k)} \in \mathbb{R}^{n} \) are the modulation parameters from the auxiliary controller \( f^{(k)}_{c} \), \ie,
\( ( \boldsymbol{\gamma}^{(:,k)},\boldsymbol{\beta}^{(:,k)} ) = f^{(k)}_{c}(\mathbf{z}) \),
such that \( f^{(k)}_{c} \) is a neural network and \( \mathbf{z} \) is the input to the controller \( f^{(k)}_{c} \). 
Note that both Equation~\ref{eq:film} and Equation~\ref{eq:malimo} linearly modulate intermediate representations, but whereas FiLM uses a single controller, MALiMo uses multiple controllers.

\paragraph{MALiMo for Audio Question Answering.}
We instantiate MALiMo for AQA as follows.
The model comprises a convolutional network \( f_{x} \) and two auxiliary controllers, as in Figure~\ref{fig:malimo}.
The first auxiliary controller \( f^{(1)}_{c} \) ingests the question to produce parameters \( ( \boldsymbol{\gamma}^{(:,1)}, \boldsymbol{\beta}^{(:,1)} ) \) that modulate the convolutional network.
The second auxiliary controller \( f^{(2)}_{c} \) ingests a subsampled version of the processed audio to produce another set of parameters \( ( \boldsymbol{\gamma}^{(:,2)}, \boldsymbol{\beta}^{(:,2)} ) \) that modulate the convolutional network as well.
Both controllers are implemented as LSTMs.
The convolutional network ingests the audio to predict an answer.

\begin{figure}[t]
  \centering
  \includegraphics[width=\columnwidth]{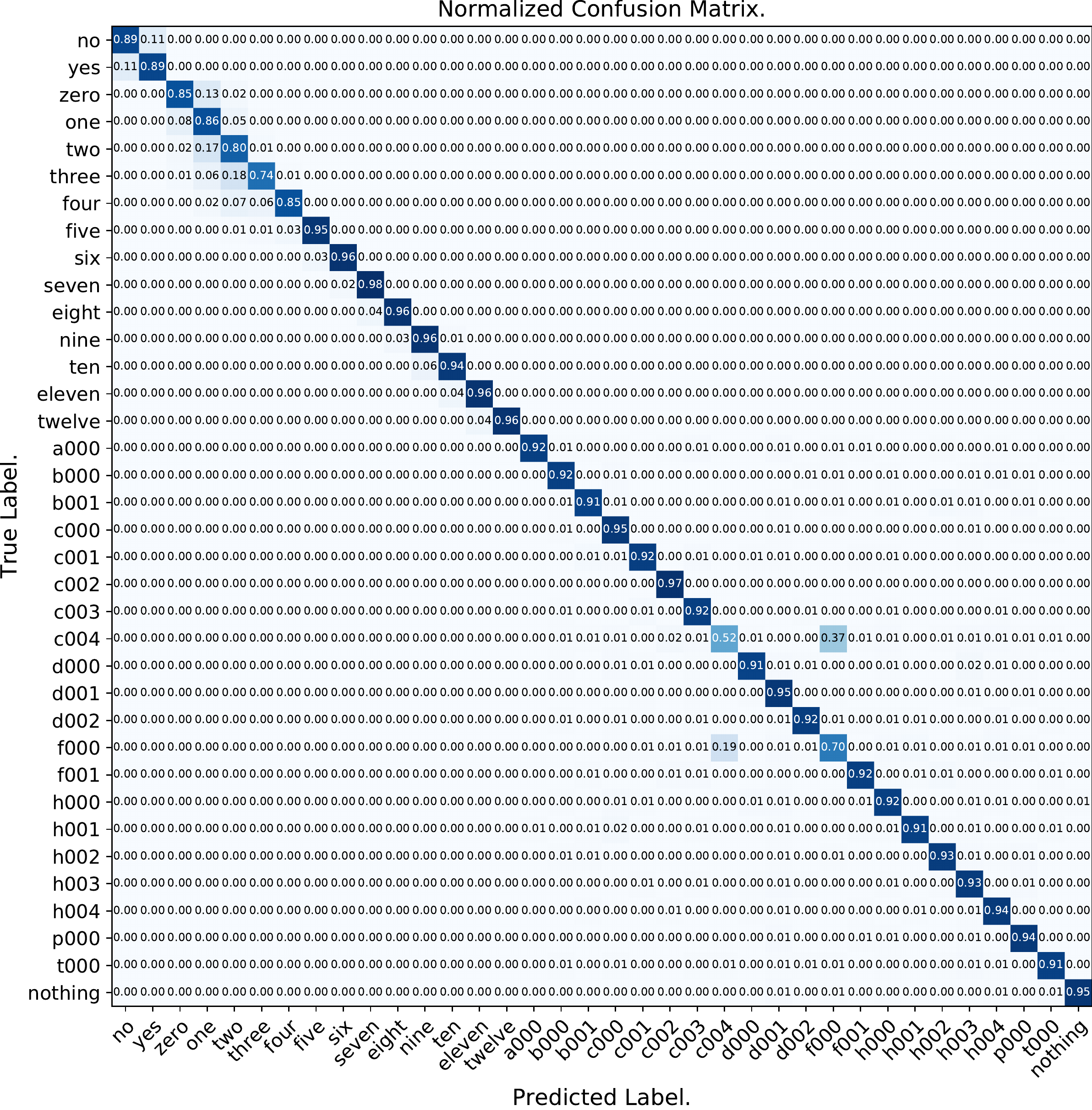}
  \caption{Normalized confusion matrix for the MALiMo (6 Blocks) model on the DAQA test set.
           Errors are predominantly due to confusion between yes and no; \( 0 \), \( 1 \), and \( 2 \); and c004 (car passing by) and f000 (fire engine passing).}%
  \label{fig:confusion}
\end{figure}

\paragraph{Implementation Details.}
Audio is converted to MFSCs, which are processed via three convolutional blocks similar to those in VGG~\cite{Simonyan2015}, such that each block comprises two convolution layers with BatchNorm and ReLUs, and a max pooling layer after the second convolution layer.
The convolution layers in the first convolutional block have \( 32 \) filters and the number of filters in the convolution layers double in each subsequent convolutional block.
The first convolution layer has filters of size \( 3 \times 12 \) with stride \( 1 \times 9 \), whereas all remaining convolution layers have filters of size \(3 \times 3\) with stride \( 1 \).
The max pooling layers have a \( 2 \times 2 \) window with stride \( 2 \).

The first auxiliary controller \( f^{(1)}_{c} \) comprises a two-layered LSTM, with \( 512 \) units in each layer, that encodes \( 256 \)-dimensional word embeddings of each word in the question into a fixed-sized representation of the question.
This representation is mapped to the first set of modulation parameters \( ( \boldsymbol{\gamma}^{(:,1)}, \boldsymbol{\beta}^{(:,1)} ) \) via a linear layer.

The second auxiliary controller \( f^{(2)}_{c} \) comprises a two-layered LSTM, with \( 512 \) units in each layer, that encodes the representation produced by the three convolutional blocks described above, subsampled using a mean pooling layer with a \( 8 \times 8 \) window with stride \( 8 \), into a fixed-sized representation.
This representation is mapped to the second set of modulation parameters \( ( \boldsymbol{\gamma}^{(:,2)}, \boldsymbol{\beta}^{(:,2)} ) \) via a linear layer.

A number of modulated convolutional MALiMo blocks ingest the output of the three convolutional blocks described above to produce a representation modulated by both auxiliary controllers \( f^{(1)}_{c} \) and \( f^{(2)}_{c} \).
Each MALiMo block comprises a stack of FiLM layers, where each FiLM layer is composed of a convolution layer with \( 128 \) \( 3 \times 3 \) filters and ReLUs, followed by another convolution layer with \( 128 \) \( 3 \times 3 \) filters, BatchNorm (with no affine transformation), feature-wise linear modulation, and ReLUs, and a residual connection from the output of the first ReLUs to output of the second ReLUs, as illustrated in Figure~\ref{fig:malimo}.
Two coordinate feature maps indicating relative time and frequency positions, scaled from \( -1 \) to \( 1 \), were concatenated to the input in each modulated block.

A convolution layer with \( 512 \) \( 1 \times 1 \) filters followed by global average pooling across each channel is applied to the output of the final modulated block and the output is fed to a fully connected neural network with a single hidden layer of \( 1024 \) units with ReLUs to predict an answer.

The experimental setup and all training details are identical to those described in Section~\ref{sec:baselines}.

\begin{figure}[t]
  \centering
  \scriptsize
  \includegraphics[width=\columnwidth]{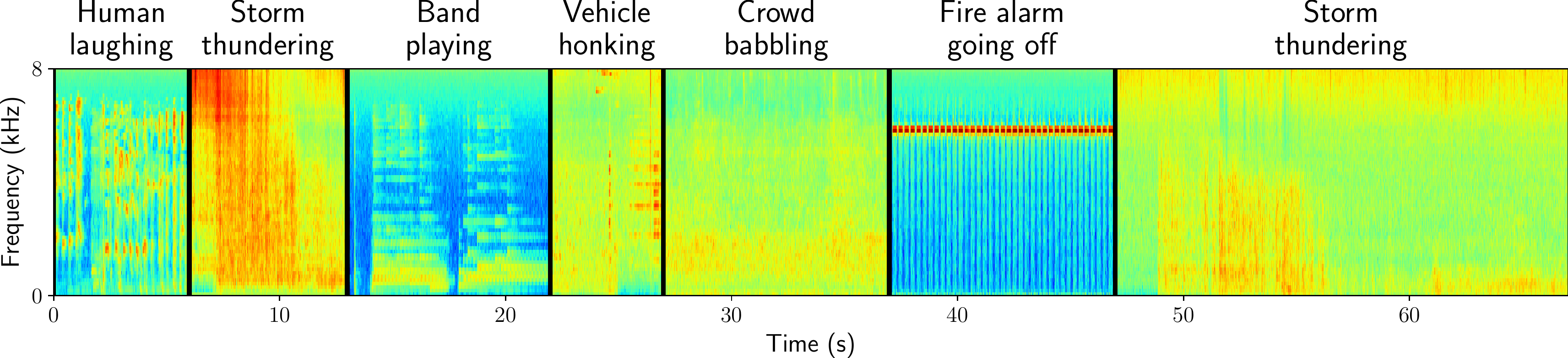}
  Hearing the second sound event did it sound like a fire engine passing by? \textbf{no} \\
  \includegraphics[width=\columnwidth]{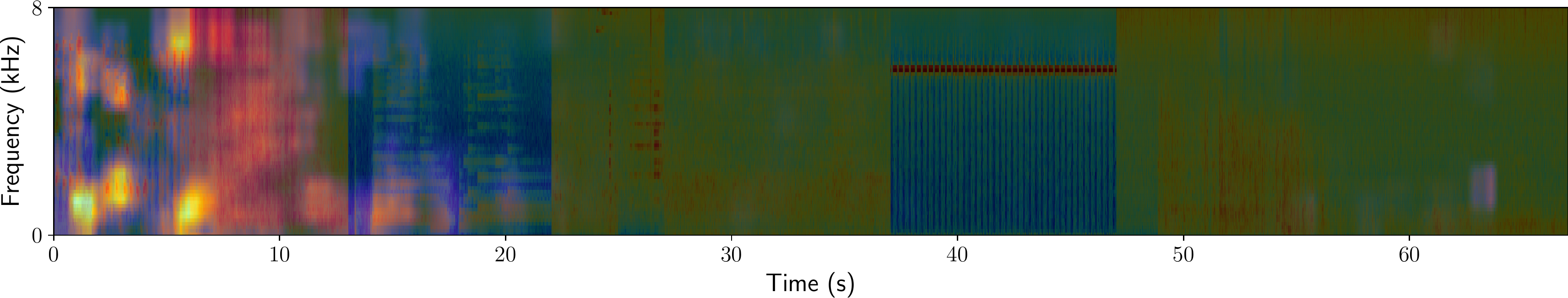}
  Was there a fire truck passing by just before the alarm going off? \textbf{no} \\
  \includegraphics[width=\columnwidth]{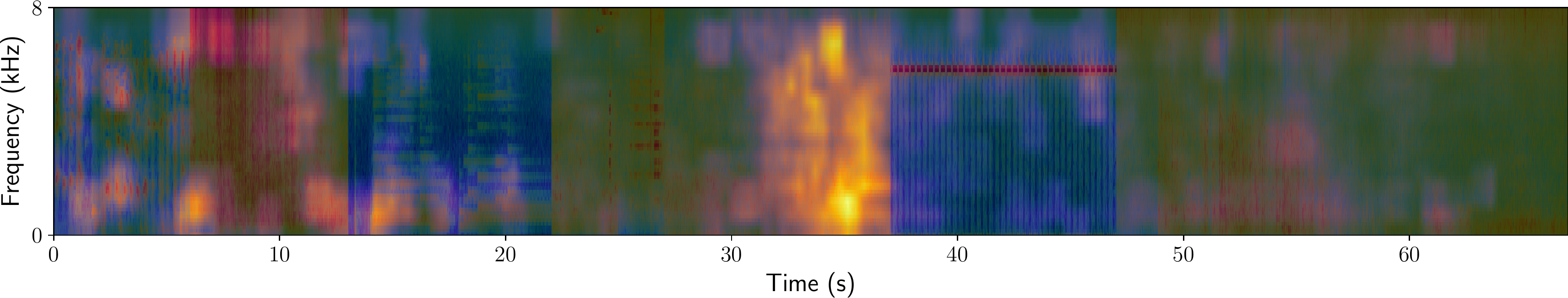}
  Was the number of times a crowd babbling less than the number of times a storm thundering? \textbf{yes} \\
  \includegraphics[width=\columnwidth]{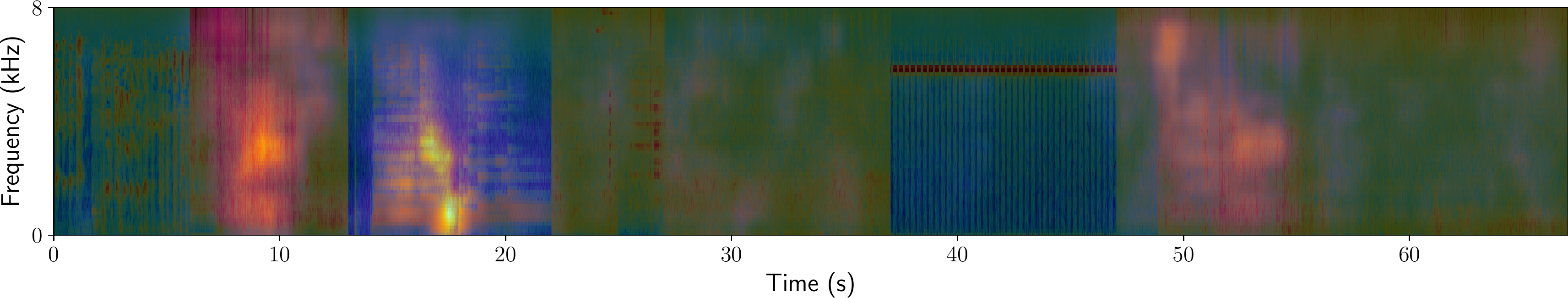}
  How many sounds after the first sound were there? \textbf{six} \\
  \includegraphics[width=\columnwidth]{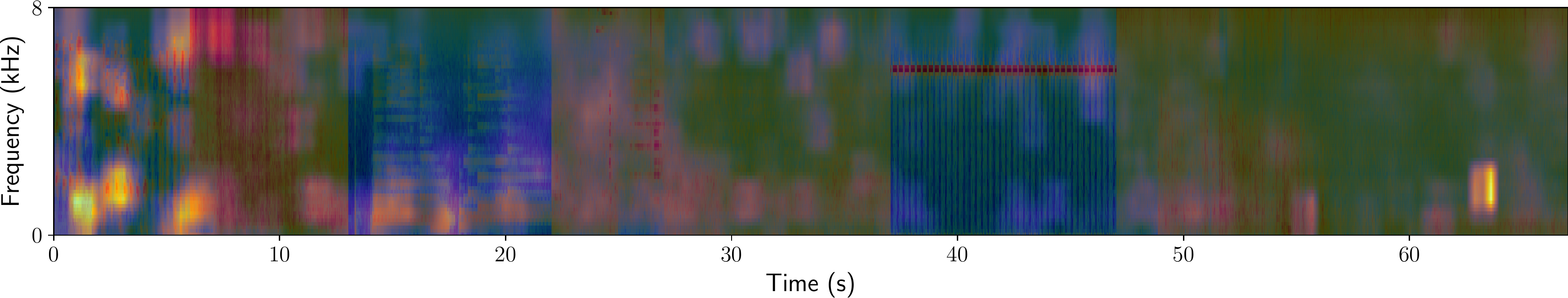}
  \caption{Saliency visualizations to highlight where MALiMo (6 blocks) attends to when answering a question about the audio.
  The first spectrogram is the time frequency representation of the audio input annotated for clarity.
  The following spectrograms are the same spectrogram superimposed with the norms of the gradient of the predicted answer scores with respect to the final convolution layer in the main module before the classifier, upsampled to match the spectrogram size, for four different questions and answers on the same audio input.
  The saliency visualizations show that the model learned to attend to the times and events of interest when performing temporal reasoning.
  Examples are from the validation set.
  }
  \label{fig:saliency}
\end{figure}

\paragraph{Results.}
MALiMo is evaluated with varying numbers of modulated blocks as shown in Table~\ref{table:results}.
MALiMo significantly outperforms all variants of FiLM and all other baselines.

Particularly, in comparison to the best FiLM model (FiLM-512; 12 layers), the best MALiMo model (6 Blocks) contains a larger number of parameters (13.20M vs 9.79M) due to the additional auxiliary controller but a similar number of layers\footnote{A MALiMo model contains roughly the same number of layers (subtracting the additional auxiliary controller) as a FiLM model with double the number of layers (\eg~FiLM (6 layers) and MALiMo (3 Blocks)), since a MALiMo block is composed of a stack of two FiLM layers.} and approximately equivalent training time, with a significant improvement in classification accuracy (88.86\% vs 78.33\%).
Doubling the size of the auxiliary controller in FiLM (FiLM-1024) to account for the capacity of two auxiliary controllers in MALiMo does not result in an improvement in performance affirming that the improvement in classification accuracy is not simply due to the additional capacity in the model.

The confusion matrix for the MALiMo (6 Blocks) model on the DAQA test set is depicted in Figure~\ref{fig:confusion}, where we observe that most errors are predominantly due to confusion between yes and no; integers: \( 0 \), \( 1 \), and \( 2 \); and events: c004 (car passing by) and f000 (fire engine passing by).

\section{Discussion and Future Work}%
\label{sec:discussion}

\paragraph{Saliency Visualization.}
Saliency visualizations in Figure~\ref{fig:saliency} are used to provide insight into where MALiMo attends to when answering a question about the audio.
Precisely, the first spectrogram is the time frequency representation of an audio input from the validation set annotated for clarity, whereas the following spectrograms are the same spectrogram superimposed with the norms of the gradient of the predicted answer scores with respect to the final convolution layer in the main module before the classifier, upsampled to match the spectrogram size, for four different questions and answers on the same audio input.
The saliency visualizations show that the model learned to attend to the times and events of interest when performing temporal reasoning.
For instance, in the second question in Figure~\ref{fig:saliency} \textit{Was there a fire truck passing by just before the alarm going off?}, the model attends mostly to the segment \textit{just before the fire alarm going off} in the spectrogram to predict the correct answer \textit{no}.

\paragraph{Temporal Reasoning.}
The convolutional network in MALiMo ingests the audio and is modulated using two auxiliary controllers: one ingests the question text, and another operates on subsampled audio features, to predict an answer.
The latter controller is crucial for relational and temporal reasoning skills as it conditions the processing in the convolutional network on the audio, which facilitates dealing with variable-sized audio and a variable number of sound events.
Figure~\ref{fig:templates} plots the classification accuracy per template for FiLM (FiLM-512; 12 layers) and MALiMo (6 blocks).
It can be seen that the majority of improvement in MALiMo over FiLM is in the templates that require temporal reasoning, \eg~contain prepositions: before / after, or ordinals, highlighted in gray in Figure~\ref{fig:templates}.

\begin{table}[t]
  \renewcommand{\arraystretch}{1.1}
  \caption{
    Audio question answering performance on the DAQA test set using the low resource training set.
    \# Blocks indicates the number of MALiMo blocks. The corresponding \# Params is listed in Table~\ref{table:results}.
    A significant drop in classification accuracy is evident compared with the same models trained using the full training set.
  }%
  \label{table:lowres}
  \centering
  \begin{tabular}{c | ccccc | c}
    \toprule
    \# Blocks        & Exist & Query & Count & Comp & CompInt & All (\%) \\
		\midrule
    1         & 58.12 & 23.63 & 36.14 & 52.31 & 52.70 & 44.93 \\
    2         & 55.94 & 13.26 & 37.40 & 51.17 & 52.88 & 42.67 \\
    3         & 52.53 & 11.93 & 34.60 & 53.97 & 50.94 & 41.79 \\
    4         & 55.53 & 13.38 & 36.87 & 50.65 & 52.29 & 42.28 \\
    5         & 55.09 & 12.82 & 34.16 & 51.10 & 52.73 & 41.61 \\
    6         & 54.35 & 11.97 & 35.06 & 51.74 & 51.25 & 41.63 \\
    \bottomrule
  \end{tabular}
\end{table}

\paragraph{Low Resource Audio Question Answering.}
All models presented so far were trained using all \( 80,000 \) audio clips and \( 399,924 \) questions and answers in the DAQA training set, which roughly corresponds to \( 5 \) (question, answer) pairs per audio clip.
A more challenging setting is to reason and generalize using a smaller training set.
To this end, we propose a new smaller training set, called low resource training set, that comprises \( 80,000 \) audio clips and \( 80,000 \) questions and answers, which corresponds to a (question, answer) pair per audio clip.
In both cases, the validation and test sets are identical to those described in Section~\ref{sec:daqa}.

We train MALiMo using the low resource training set and evaluate the models using the DAQA test set.
The experimental setup and all training details are identical to those described in Section~\ref{sec:baselines}.
Table~\ref{table:lowres} lists the classification accuracy on the DAQA test set for a number of MALiMo models.
A significant drop in classification accuracy is evident compared with the same models trained using the full training set listed in Table~\ref{table:results}.
These results indicate that developing models that can reason and generalize from a relatively small dataset remains a challenging task for future work.

\paragraph{Selective Computation.}
The models studied herein are capable of dealing with variable-length audio up to approximately 3 mins;
however, the models must process the entire audio clip to produce an answer, which may not be efficient for cases such as, \textit{What was the first sound?} or \textit{What was the last sound?}.
Thus, models that can adaptively navigate the audio clip, without necessarily processing the entire clip, would be a desirable avenue for future work~\cite{Mnih2014, Yeung2016}.

\section{Conclusion}%
\label{sec:conclusion}

We demonstrate that the AQA task can be used to study the temporal reasoning abilities of machine learning models.
We introduce the DAQA dataset that comprises audio sequences of natural sound events and programmatically generated questions and answers that probe various aspects of temporal reasoning.
We show that several recent state-of-the-art methods for VQA adapted to the AQA task perform poorly, particularly on questions that require in-depth temporal reasoning.
We propose a new model, MALiMo, that utilizes auxiliary controllers to modulate the intermediate representations of a convolutional network processing the audio conditioned on the question and the audio itself.
Experiments on DAQA show that MALiMo significantly outperforms other competing approaches, and that it leads to the largest improvements on questions that require in-depth temporal reasoning.
Avenues for future work include learning to reason and generalize from small datasets and developing models that can process only relevant segments of the audio as opposed to the entire audio clip.
We envisage DAQA to foster research on AQA and temporal reasoning and MALiMo a step towards models for AQA.

\ifCLASSOPTIONcaptionsoff
  \newpage
\fi

\bibliography{references}
\bibliographystyle{IEEEtran}

\newpage

\appendices

\section{Diagnostic Audio Question Answering (DAQA) Examples}

\begin{figure*}
\centering
\resizebox{.8\textwidth}{!}{
\begin{tabular}{c}
\toprule
\includegraphics[width=\textwidth]{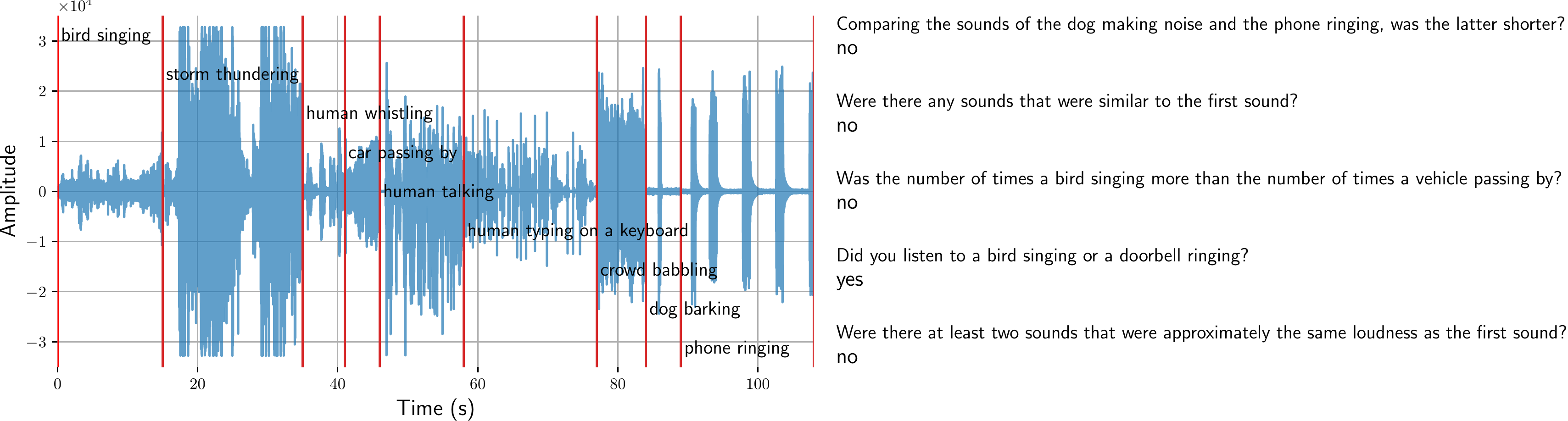} \\
\midrule
\includegraphics[width=\textwidth]{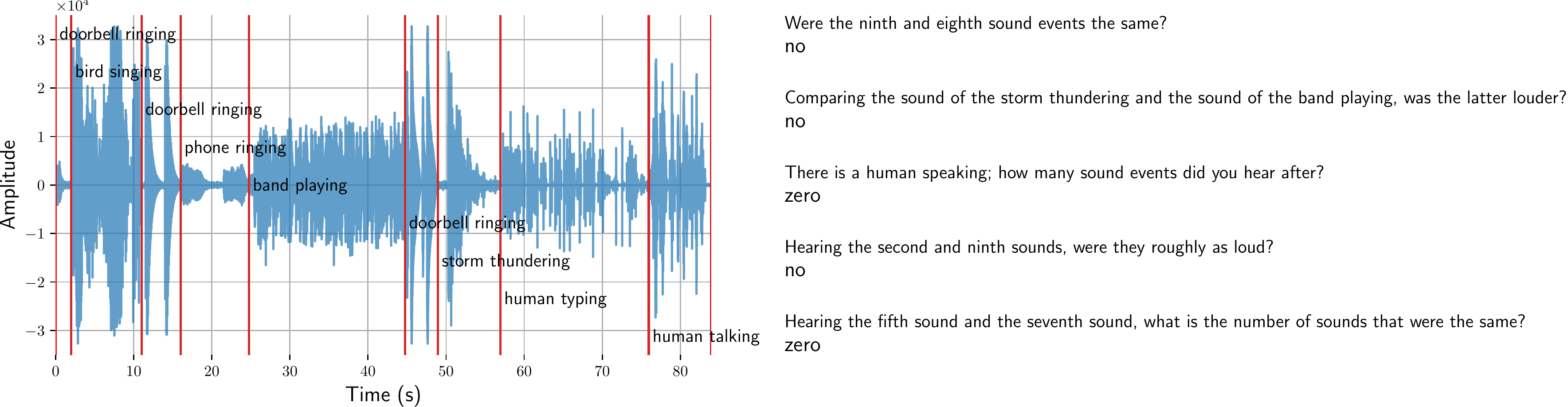} \\
\midrule
\includegraphics[width=\textwidth]{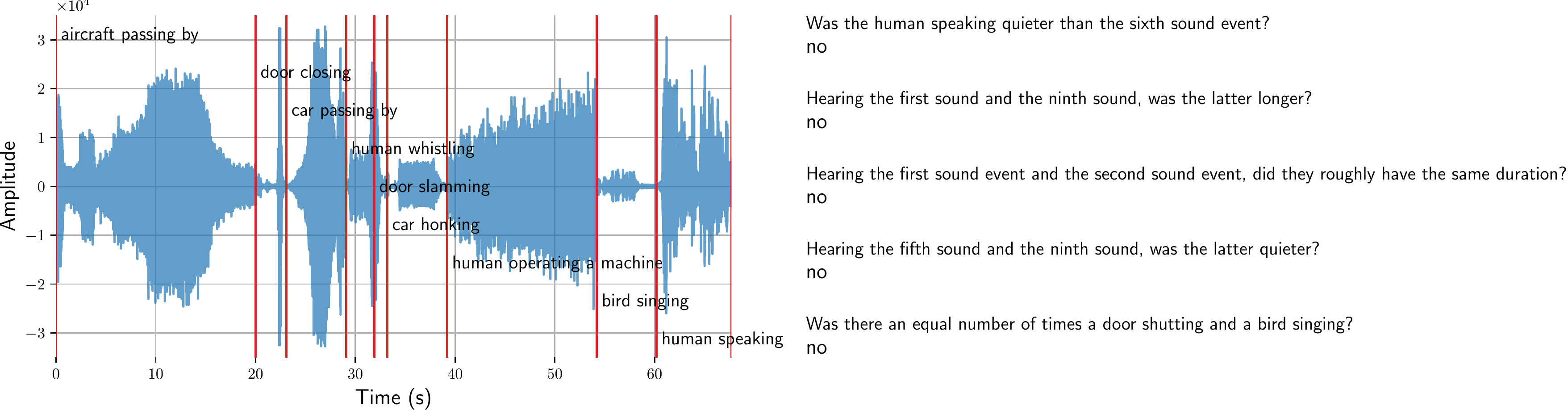} \\
\midrule
\includegraphics[width=\textwidth]{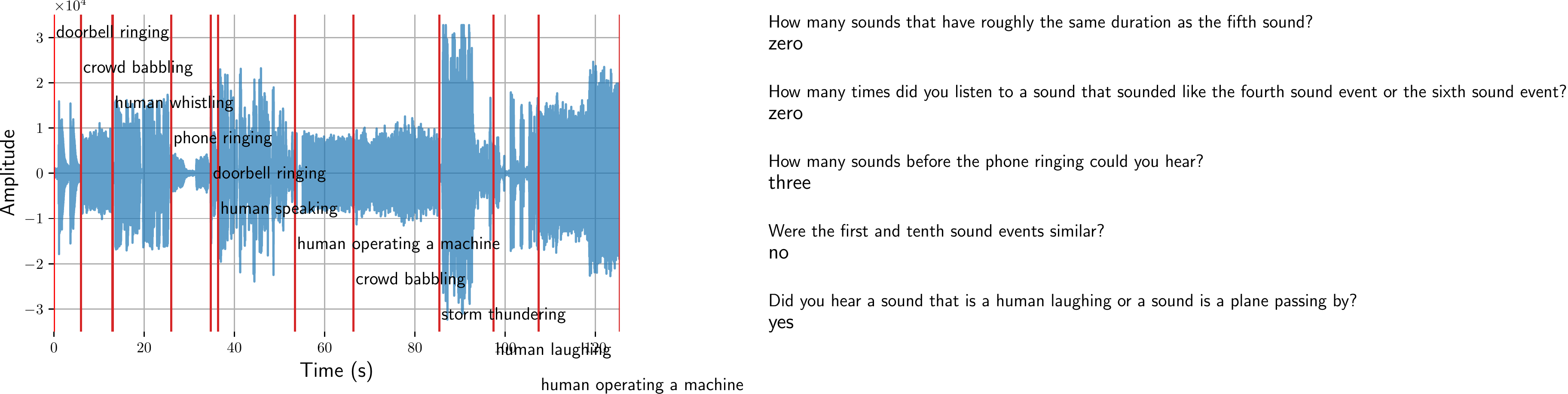} \\
\midrule
\includegraphics[width=\textwidth]{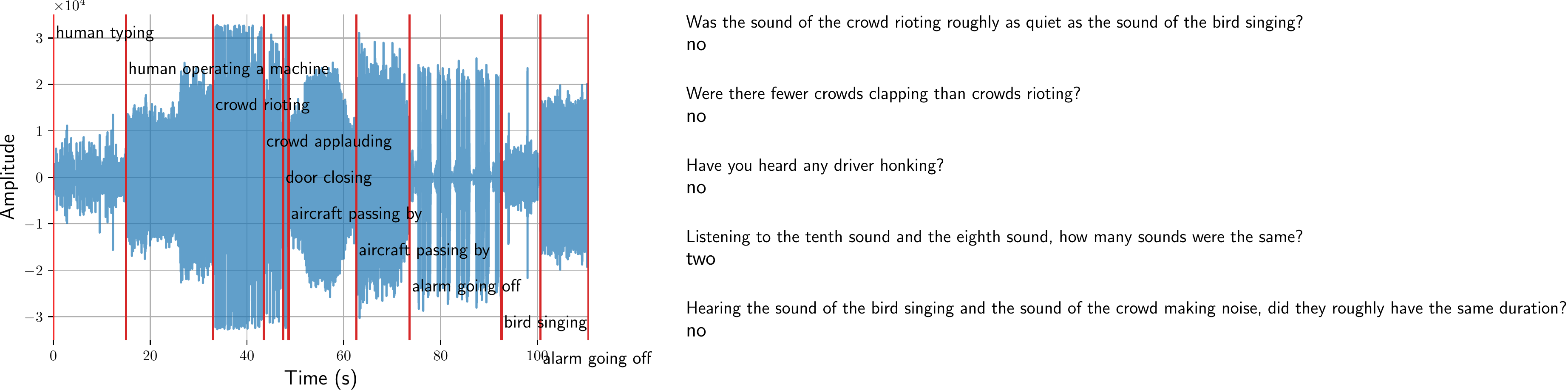} \\
\midrule
\includegraphics[width=\textwidth]{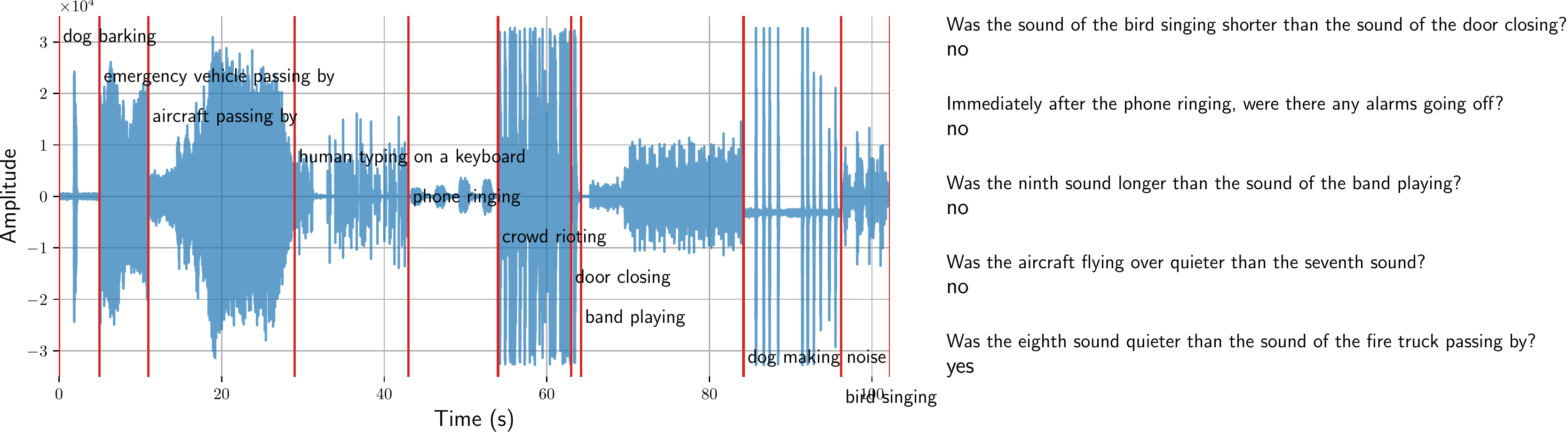} \\
\bottomrule
\end{tabular}
}
\caption{Audio clips composed of natural sound events (left) and programmatically generated questions and answers about those audio clips that test various aspects of temporal reasoning (right).}
\label{fig:examples}
\end{figure*}

Figure~\ref{fig:examples} is a number of random examples from the training set of DAQA.

\end{document}